\documentclass[journal,a4paper]{IEEEtran}
\usepackage{graphicx}
\usepackage{balance}
\usepackage{color}
\usepackage{ulem}				
\usepackage{listings}				
\begin{document}

\title{Fault Tolerance in Distributed Neural Computing}

\author{Anton Kulakov,
Mark~Zwoli\'nski, and
Jeff Reeve
\thanks{Mark Zwolinski is with Electronics and Computer Science, University of Southampton, Southampton, SO17 IBJ, UK
e-mail: mz@ecs.soton.ac.uk}}

\maketitle

\begin{abstract}
With the increasing complexity of computing systems, complete hardware reliability can no longer be guaranteed. We need, however, to ensure overall system reliability.
One of the most important features of artificial neural networks is their intrinsic fault-tolerance. The aim of this work is to investigate whether such networks have features that can be applied to wider computational systems. 
This paper presents an analysis, in both the learning and operational phases, of
a distributed feed-forward neural network with decentralised event-driven time management, which is insensitive to intermittent faults caused by unreliable communication or faulty hardware components. The learning rules used in the model are local in space and time, which allows efficient scalable distributed implementation.  We investigate the overhead caused by injected faults and analyse  the sensitivity to limited failures in the computational hardware in different areas of the network.
\end{abstract}



\section{Introduction}
The inevitable demand for ever more computational capability drives the creation of ever larger parallel distributed machines (e.g. the K computer has 705,024 cores \cite{top500} 
), so that although the mean time to failure (MTTF) for the individual components can be very high (up to $10^{6}$ hours \cite{Mukherjee:2003}), the large number of components will inevitably lead to frequent failures -- on average once every one and a half hours.
Failures can also be caused by the fact that in large multi-processor systems the arrival of communication messages is not guaranteed or they can arrive late \cite{SpiNNaker:2010}. This requires new solutions for fault tolerance to allow the next generation of extreme-scale massively parallel computers to be used at their full-capacity. 

In recent years, it has been suggested that neural computing offers a model of fault-tolerant computing. In biological brains,  neurons die without apparent loss of functionality of the whole system and by analogy, this principle has been applied to neural simulation engines (SpiNNaker \cite{SpiNNaker}, BlueBrain \cite{BlueBrain}). For instance, SpiNNaker will employ 50,000 chips (with 20 slow -- 200 MHz -- processors per chip) connected over a fast network (1 Gbps) and is based on a model of communication-centric computation, in contrast to conventional, calculation-centric computers with very fast processors (2 GHz) over not-so-fast networks. In order to maintain high communication speed and avoid deadlocks, a packet-dropping mechanism is used when a packet cannot be forwarded \cite{Plana:2009}. 

Artificial neural networks have been inspired by studies of the brain structure, where information is processed in a parallel and distributed way. Commonly used conventional sequential computing systems utilize one or a few sparsely-interconnected, high performance processing units. Neural networks, in contrast, employ a large number of highly interconnected, very simple processing elements, where the computational power of the model comes above all from the interaction of all its units. 

The motivation for this work 
is to determine whether neural computing can be used as a paradigm for reliable systems running on unreliable hardware.
We examine the fault-tolerant characteristics of parallel distributed processing networks with a feed-forward structure, 
in order to understand how the required fault tolerance can be achieved on systems with unreliable communications. The work investigates neural network performance under damage conditions and dynamics of weight change in a representative task.



This paper is structured as follows. We first review related work in the field and outline several techniques to assure the fault tolerant behavior of neural networks. Then we define the key terms and concepts for general and comparable results as well as discussing the network structure and training techniques used in the experiments. Next we relate these findings to an analysis of the network's structure. The implications of the findings for fault tolerance and its improvement are then discussed, before concluding.

\section{Related Work}
\subsection{Fault tolerance in neural networks}

Due to the multiplicity of individual units, neural networks contain more processing elements than is necessary to solve a problem. Moore \cite{Moore:1989} argued that because of the large number of components, neural computers would need to be designed with high quality components. In contrast, Mozer's work \cite{Mozer:1989} has shown that units may be simply removed from a network without damaging its performance. The loss of a few units would be unlikely to cause any noticeable decrease in accuracy of the overall performance in a large system. Tai in \cite{Tai:1990} showed that losses of up to 40\% can be tolerated in the Hopfield model. 

This leads to the conclusion that due to the inherent overall fault tolerance, non-critical components within the neural network system need not be particularly reliable, as has been proposed by Belfore and Johnson \cite{Belfore:1989}. Based on this conclusion, Chiu et al \cite{Chiu:1993} proposed an algorithm to improve both the efficiency and fault-tolerance of a multi-layer network. In their approach a unique measure of neuron relevance is used, according to which the least significant neurons are eliminated, whereas the most significant ones are duplicated. 

Carter and Segee \cite{Carter:1991} empirically showed that multilayer networks do not significantly reduce the level of tolerance after pruning. However, they also pointed out that this is not always the case. Despite inherent fault tolerance being provided by the distributed processing architecture, neural networks are not always tolerant of the loss of processing elements \cite{Carter:1990}. Their conclusion is that often, instead of catastrophic failure under the influence of faults or noise affecting inputs and internal components, it is most likely that network performance will degrade gracefully.

Furthermore, Segee et al showed in \cite{Segee:1994} that fault tolerance is influenced by the training algorithm used and even the initial state of the network. The implication is that if the number of network processing elements can be made large, then fault tolerance increases in the network automatically by virtue of the gross similarity with biological neural networks. The idea is that fault tolerance in a neural network is directly related to the redundancy introduced because of ``spare capacity'', when the complexity of the problem is less than the computational capacity of the network. Nijhuis et al. \cite{Nijhuis:1990} came to a similar conclusion, stating that fault tolerant behavior is not always self-evident but must be assured by an appropriate training scheme. Furthermore, \cite{Damper:1993} presents a procedure to build fault tolerant neural networks by replicating the hidden units. An analytical derivation of the minimum redundancy required in order to tolerate all possible single faults is presented in \cite{Phatak:1995}.

On the other hand, von Seelen and Mallot \cite{Seelen:1989} discuss whether indeed a neural network's reliability is caused by redundancy, both in terms of fault tolerance and graceful degradation. They assume that a neural network uses all of its resources to balance between computational accuracy and computation time. Thus redundancy is not identical to reserve capacity and neural networks utilize available resources to the full.

Taking these sometimes contradictory findings into account, 
we have examined the fault tolerance of neural networks in both the training and operational phases, so as to be able to evaluate the system reliability when such networks are implemented on massively parallel hardware. Moreover, the published literature considers failures in the \textit{computational} units; we are just as concerned with failures in the \textit{communication} links.

\subsection{Concepts of Reliability}
The field of reliable and fault-tolerant computation is very wide, embracing many different architectural and operational features of neural network systems as well as several conceptual viewpoints. However, in the literature an inconsistent and often inaccurate use of key terms and concepts can be found, which can cause confusion and uncertainty
. A categorization of the causes of failures affecting neural network reliability must be developed in order to omit non-precise terms and obtain general and comparable results. This section addresses the problem and defines key terms and concepts, used further in the paper.

First of all, to describe the reaction of the network performance to faults, the terms \textit{graceful degradation} and \textit{fault-tolerance} are often used. Unlike conventional computers, a neural system  is often not adversely affected by faults or noise in internal components. Instead of failing catastrophically, the system continues delivering acceptable, although possibly reduced performance. Computational accuracy is allowed to degrade in a controlled manner as the fault severity increases. Such a low sensitivity to occurring faults instead of a complete failure is known as \textit{graceful degradation} \cite{Bolt:1991}.

In contrast to graceful degradation, \textit{fault tolerance} is the property that guarantees the \textit{proper} operation of the system in the event of a failure (or several failures) within some of its components. It describes the robustness of the network function in the presence of degradation in the computational elements, such as broken connections or erroneously functioning processing elements.


A common misunderstanding is that of confusing fault tolerance with robustness to noisy inputs. \textit{Fault tolerance} is the ability of a system to continue to perform to specification in the presence of hardware faults, such as broken connections, connections with an erroneous weight, or neurons with inaccurate outputs. On the other hand, a system that continues coping with input noise and operates correctly despite errors in its inputs is termed a \textit{robust} system. However, it is fault-tolerance, rather than robustness, that is associated with sensitivity to internal noise. Nijhuis in \cite{Nijhuis:1990} refers to fault tolerance as \textit{hardware fault-tolerance} and correspondingly to robust systems as \textit{data fault-tolerant} systems. In this paper, we focus exclusively on hardware fault-tolerance, which describes a system's sensitivity to faults that result in perturbations in network parameters or topology, but does not refer to noisy or partial input data. In fact, we believe that the sensitivity of a system to noisy inputs is both inappropriate and inconsistent for defining a fault-tolerant system.

Another area of potential confusion is the stage at which errors start occurring. Carter in \cite{Carter:1990} distinguishes between two types of fault-tolerance in neural networks: an \textit{operational} and a \textit{learning} fault-tolerance. The sensitivity of network performance to permanent or transient faults occurring at the learning stage, is referred to as \textit{learning} fault-tolerance. Whereas the \textit{operational} fault-tolerance deals with the sensitivity of network performance to faults presented after learning has been accomplished in a fault-free environment. In this paper we pay primary attention to learning fault-tolerance.

\section{Fault Simulation}
Neural networks are often treated as black-box systems. Measuring the degree of failure is based on the results at the output units for presented input data. We investigate what level of fault-tolerance the neural network can achieve given the faults that might occur during both learning and operation phases. In the case of a fault occurring during the learning phase, we consider how much longer it will take to train the network and how fault-tolerant the final version of the network will be. To address all of these issues, artificial faults can be introduced into the system. This section discusses the possible ways of introducing faults.

\subsection{Fault-tolerance Analysis}
In order to achieve a suitable analytical model of the reliability of a neural network system, a definition of various failure modes and their impacts on the system is required. This would lead to a firm foundation for further investigations of the amount of fault tolerance exhibited by a neural network. However, taking into account the high level of inter-dependence of elements in the system on each other, this task is extremely complicated. 

Due to the difficulty in defining the effect of an individual unit or connection on the overall reliability of an entire system, we use empirical investigation. We consider a type of fault that is admittedly severe and correspond to the highest failure rates. It is often referred as ``loss of weight'' fault and occur in the case of open-circuits, \cite{Carter:1990}. Some authors refer to this fault as ``stuck-at-0'', e.g. \cite{Bolt:1991}. A classical example of an open-circuit fault can be imagined as a damaged neuron or connection, which can be related in biological terms to the continual loss of synapses in the brain. In simulation it is implemented by setting the selected weight to zero.

\subsection{Fault Injection Technique}

During each simulation, faults are probabilistically introduced and the degree of failure is evaluated according to some measure. The measure of reliability from many experiments can be plotted against the number of introduced faults injected into the system. The plot indicates the way the neural network model behaves depending upon the generic nature and the faults occurrence rate. Different plots can be compared and contrasted in order to judge the system's sensitivity to different types and locations of faults. This facilitates evaluation of fault-tolerance when the type of fault and the rate of occurrence are known.

The fault injection technique is convenient for indicating the isolated effects of individual faults. However it is impractical for evaluating the impact of multiple faults as their effects combine and are not independent. It is not realistic simply to add the impacts of single faults in order to imitate the effect of multiple faults due to non-predictable correlations between them. This complicates accurate prediction of the effect of all faults occurring together over a period of time in real use. Also another scenario is possible: that a system maintains an adequate performance despite a limited injection of faults. However, after a certain fault threshold is reached, the system may abruptly reduce its performance, which can lead to a total failure.

Another complication is caused by the temporary nature of many faults, often called ``transient'' and ``intermittent'' faults. Transient faults are non-recurring and intermittent faults recur at, usually irregular, intervals. These faults are caused by several contributing factors, some of which may be effectively random, which occur simultaneously. The more complex the system or mechanism involved, the greater the likelihood of an intermittent fault. An estimation of the impact of temporal faults is unreliable and thus is out of the scope of this investigation. 

In conclusion, a fault injection method is useful for gaining a very basic indication of the reliability of a neural network system, though it may identify especially critical areas of a neural network which can then be protected against possible faults in any implementation.

\section{Experimental Approach}
In terms of fault tolerance, some neural networks and some training algorithms are better than others. If nothing is done to control the proper operation of the system in the event of a fault during training, the fault tolerance of the final network may be very random, depending on the problem, the chosen architecture, the data representation, and the learning examples. In this section we discuss the network structure, training technique and teaching procedure used in the experiment.

\subsection{Network Structure and Connectivity}
Training physically distributed neural networks with no shared memories and decentralized event-driven time management has always been a challenging issue. On one side, the absence of shared memory excludes data contentions. But at the same time there is no conscious control over the spike generation, emitting, storing, and processing. Each processor sequentially performs event processing in accordance with the temporal order of these events. Nothing ensures that events are processed in a correct order.

\begin{figure}
\begin{center}
\includegraphics[width=\columnwidth]{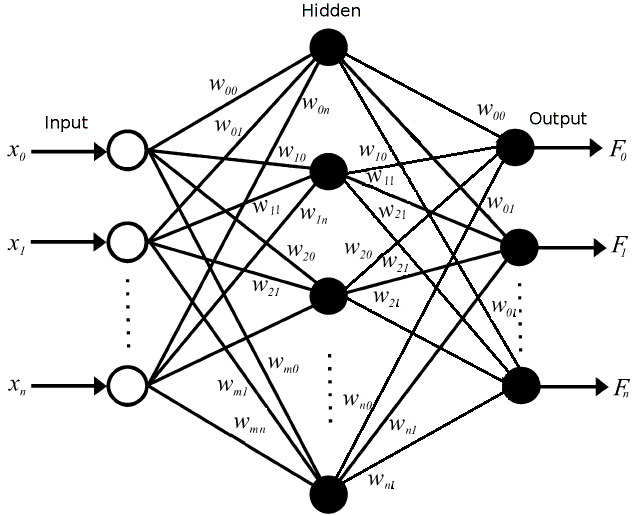}
\caption{The architecture of a feed-forward neural network with \textit{n}-input, \textit{m}-hidden layer and \textit{l}-output node.}
\label{netStruct}
\end{center}
\end{figure}

Of all the existing neural network topologies, the feed-forward neural network is probably the one most often used. This network topology consists of three or more layers: an input layer, an output layer and a number of hidden layers in between. Each layer contains a number of neurons, which are connected only with the neurons of the adjacent layers. Activity flows from input to output and the network topology contains neither cycles nor lateral connections. The input layer is present merely to increase the fan-out of the input data, whereas the hidden and output layers perform computations, as shown in Fig. \ref{netStruct}.

Commonly, the input data is represented in a binary form, corresponding either to the presence or absence of features. The actual process is based on the collective computations that are performed in the synapses and the neurons. At each synapse the incoming signal is multiplied by the corresponding connection strength, $w_{ij}$. At the neuron, these values are summed and compared to the threshold value: if the summed result exceeds the threshold value, a neuron emits a consecutive spike, otherwise it remains inactive. 

For nearly all problems, one hidden layer is sufficient. Using two hidden layers rarely improves the model, and may introduce a greater risk of convergence to a local minimum and there is no theoretical reason for using more than two hidden layers \cite{Panchal:2011}, \cite{Huang:2003}. Two hidden layers are, in principle, enough to perform any classification task \cite{Burr:1988}, including high-level abstractions (e.g. in vision, language, and other AI-level tasks) \cite{Bengio:2009}. Here, the number of hidden layers is limited to one.

In our approach, a special processor is dedicated to interacting with the environment and at the same time managing the work of other processors (e.g. network mapping, setting the synchronization barrier). The master processor only `knows' which processors are dedicated to representation of the input and output neurons. A slave does not `know' whether it contains input, hidden or output neurons. Each slave processor is idle until it receives a spike message (from another slave or the master processor). When the spike message is received, and the excitation conditions are met, a new spike messages are sent to the all known targets, determined \textit{a priori} while creating the network. 

\subsection{Network Training}
The dynamics of a neural network are determined by a rule, derived by Bosman et al \cite{Bosman:2004}. This is a form of reinforcement learning, where the active weights are either incremented or decremented by a certain weight proportion based on the binary feedback signal in accordance with Hebbian learning. The signal represents the `success' or `failure' of a given output after each attempt to associate the correct output with a particular input. The important benefit of the model is the locality of all the necessary information about the states and the properties of the involved neurons for calculating weight alterations. This allows simulation of distributed neural networks with decentralized, asynchronous time management.

A network usually has two different operating modes: \textit{training} and \textit{operation}. Identical faults are likely to have different effects during these two distinct phases. During the \textit{training stage}, weights between the neurons are adapted in order to \textit{learn} to reproduce a set of patterns, which represents the problem. The actual training stage of the neural network consists of many so-called \textit{training cycles}. During each training cycle 
the network attempts to alter its weights in such a way that the output neurons produce an expected pattern. The network is said to function when all training pairs of patterns are correctly `memorized'. 

The number of training cycles depends on various factors such as (a) the complexity of the decision regions that is caused by the data itself; (b) the network topology, that is to say the  number of layers and the number of neurons per layer; (c) the learning strategy, which consists of the choice of parameters in the training algorithm and the order of presentation of training patterns during learning; and (d) the rate of forgetting due to interference of the new learned data with that previously learned.

In the case where the excitation conditions were met first time for the current input pattern, spike messages are sent to the post-synaptic neurons. At the same time another message containing the value of the membrane potential is sent to the pre-synaptic neuron from which the last spike message  arrived, indicating that pre-synaptic neuron's spiking activity led to the excitation of the post-synaptic neuron. The pre-synaptic neurons store this information in the \textit{activated} array. If the excitation condition is not met  for the first time for the current input pattern, the neuron informs the pre-synaptic neurons about the alteration of its value of membrane potential. In the case that it receives an inhibitory signal and is not active any more, the neuron sends another message to its post-synaptic neurons to warp back the neurons's influence onto them, changing their state correspondingly.

Finally, when the output neurons produce the expected output pattern, the master sends a message, which either reinforces the weights, if the received output corresponds to that expected, or reduces (`deinforces') the weights otherwise. Two arrays (\textit{activated}  and \textit{activatedBy} ) contain all the necessary information for applying the learning rule described in \cite{Bosman:2004}. When the weights are altered, a new cycle starts.

Faults occurring during the learning stage help the network's resilience to possible damage in the operation stage because of the more evenly distributed information between its weights \cite{Chiu:1994}. This prevents the situation of random information distribution among the weights, when some connections are very influential on the network output (and thus also very sensitive to perturbations), whereas others are almost useless.

\subsection{Forgetting}
All natural cognitive systems gradually forgets previously learned information as new information is acquired. A similar effect is observed in artificial neural networks. While storing a large amount of input patterns into the network, interference of newly learned data with previously learned data occurs. It turns out that in the distributed computational nature of neural networks, the very virtue of the approach is at the same time the root cause of forgetting. 

The negative effect of the \textit{path interference} arises when several input patterns are applied to the neural network to be memorized. The active paths overlap when the strongest connection from the different input patterns point to the same intermediary neurons. As a result, the learning of something new causes forgetting of old data. The challenge is how to keep the advantages of distributed computation while avoiding the problem of catastrophic forgetting. For this it is necessary to examine the basis of the phenomenon.

\begin{figure}[!h]
\begin{center}
\includegraphics[width=\columnwidth]{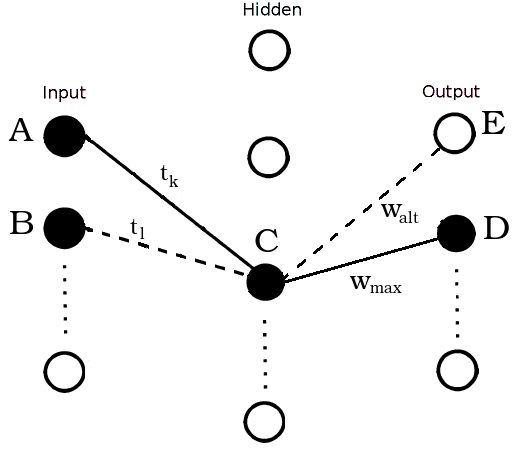}
\caption{An example of \textit{path interference} between A-C-D and B-C-E paths. Path A-C-D was formed at learning step \textit{$t_{k}$} and became the strongest one with weight efficacy \textit{$w_{max}$}, at learning step \textit{$t_{l}$} affects the formation of B-C-E path with weight efficacy \textit{$w_{alt}$}.}
\label{pathInt}
\end{center}
\end{figure}

There may be several reasons for such a situation. First of all, from the active input neuron the path of activity runs along the strongest synaptic connections to the corresponding output neurons. In certain situations an established path can be completely ``wiped out'' by an attempt to learn new data, so that the connection of the previously learned pattern is no longer the strongest. Also a competition between the active path, formed in the previous steps, and the newly forming active path can occur. Such competition often
erases or partially destroys the old path and correspondingly leads to forgetting of old data by the network.

Corresponding to each input, the most probable signal propagation will follow the associated pattern. However, when the number of input patterns or the inter-connectivity level increases, the activity paths overlap, thereby destroying each other and corrupting the output result.

The intuitive solution is a uniform distribution of active paths in the network. This can be achieved in two ways. Firstly, a negative influence of active path interference can be partially overcome by periodically shuffling the input-output pairs of patterns before feeding them into the network, as described in \cite{Kulakov:2011}. This approach assists in finding the most optimal weight values valid for all patterns and, additionally, network's fault tolerance increases due to more even weight distribution throughout the network.

\subsection{Network Operation}

During the \textit{operation stage}, weights are unalterable. The percentage of errors in recognizing the set of pre-learned patterns \textit{($x^{l},d^{l}$)}, \textit{l = 1, ..., p} will be called the \textit{global error} and is calculated in the following way. 

Let \textit{$w_{ij}$} represent the weight in the \textit{i}-th row and \textit{j}-th column of the weight matrix \textbf{W}. The \textit{n}-dimentional input pattern is set on the input neurons, propagated through the network, transformed according to the current state of the weights and the corresponding \textit{m}-dimensional output of the network \textit{$y^{l}$} is compared with the desired one \textit{$d^{l}$}. 

The error for the output neurons per  pattern is calculated according to the formula:

\begin{equation}
 Error_{l}(\textbf{W}) = \frac{1}{m}\sum_{k=1}^{m}{(d_{k}^{l} - y_{k}^{l})^{2}}.
\label{equation}
\end{equation}

Then, the global error is defined as:
\begin{equation}
 Error(\textbf{W}) = \sqrt{\frac{1}{p}\sum_{l=1}^{p}{Error_{l}(W))}}.
\label{equation2}
\end{equation}

So that when the actual output \textit{$y^{l}$} is equal to desired output \textit{$d^{l}$}, the global error equals to zero. 

\section{Simulation Results}
The distributed neural network simulator was written in the C++ language using the Message Passing Interface library (MPI) for inter-processor communication. The simulated neural network consisted of an input layer, one hidden layer and an output layer. Decentralized event-driven time management without memory sharing was applied.



\subsection{Fault Tolerance in Training}
Are all connections equally significant? To answer this question we investigated the weight alteration dynamics during the training process in order to identify the most and the least significant connections. Along with the network structure, the initial state of the weights has a large impact on the ability to learn and the resilience of a neural network. In this section we look into the optimal initial values of the weights.

\begin{figure*}
\begin{center}
\includegraphics[width=\textwidth]{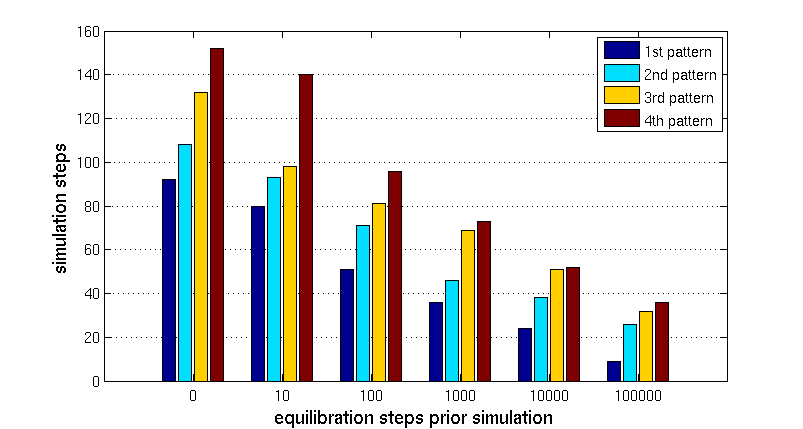}
\caption{Training efficiency dependence on the number of equilibration steps applied prior to training. During the training phase 4 different patterns were learned. The network size was 2000 neurons with 5 input and 5 output neurons and 90 \% connectivity.}
\label{equil}
\end{center}
\end{figure*}

We constructed a small network of 20 neurons (for better visibility) and recorded the alteration of each weight, plotting the measured values against each simulation step. We noticed that training efficiency directly depends on the initial state of the network: weight equilibration before the learning phase significantly improves training process. This is demonstrated in Fig. \ref{equil}. 1000 random inputs were applied to the network 10000 times before starting any measurements. For each input the ``deinforcement'' process was applied to the weights, modifying the overall distribution of their values in the network. 

\begin{figure}
\begin{center}
\includegraphics[width=\columnwidth]{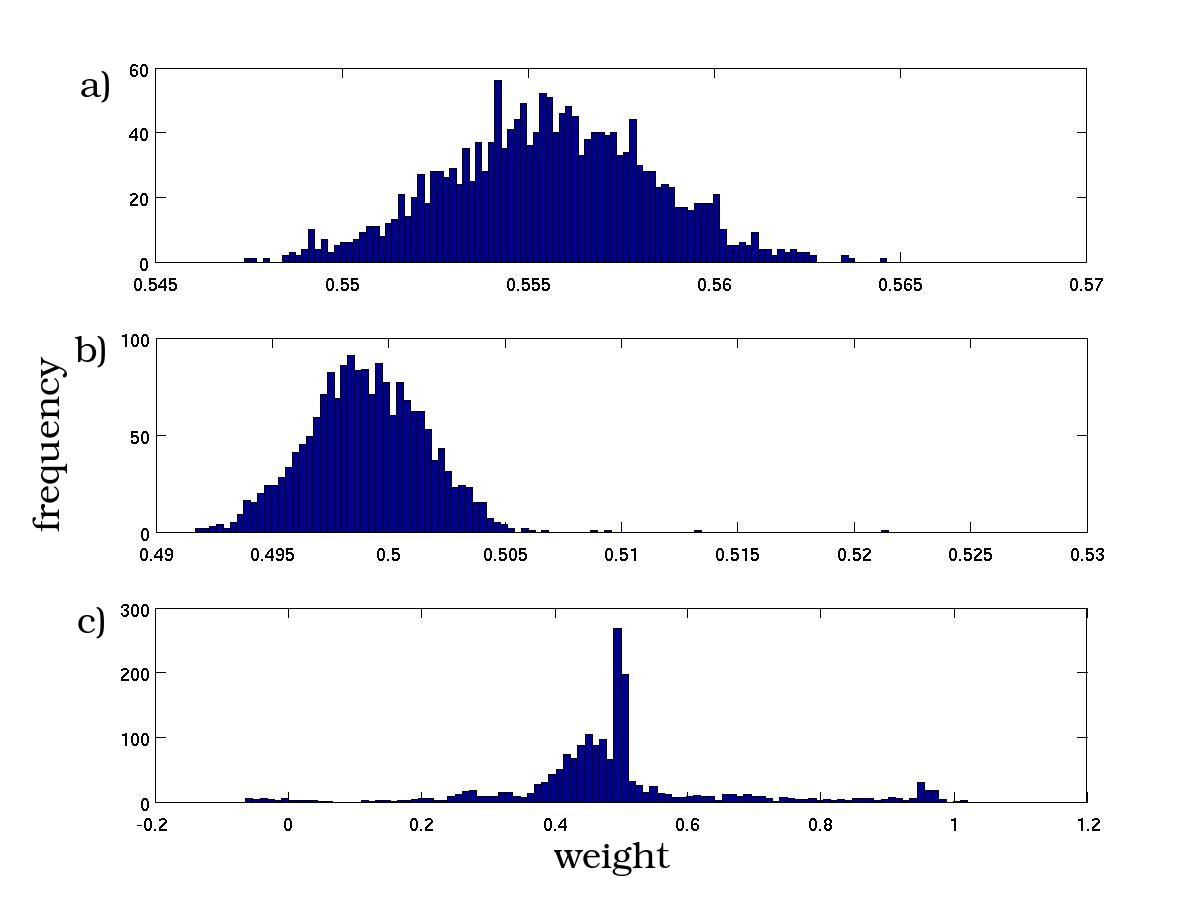}
\caption{Weight distribution of input-hidden connections at the initial phase (a), after the equilibration phase (b), and after the learning phase (c).}
\label{hidd}
\end{center}
\end{figure}

\begin{figure}
\begin{center}
\includegraphics[width=\columnwidth]{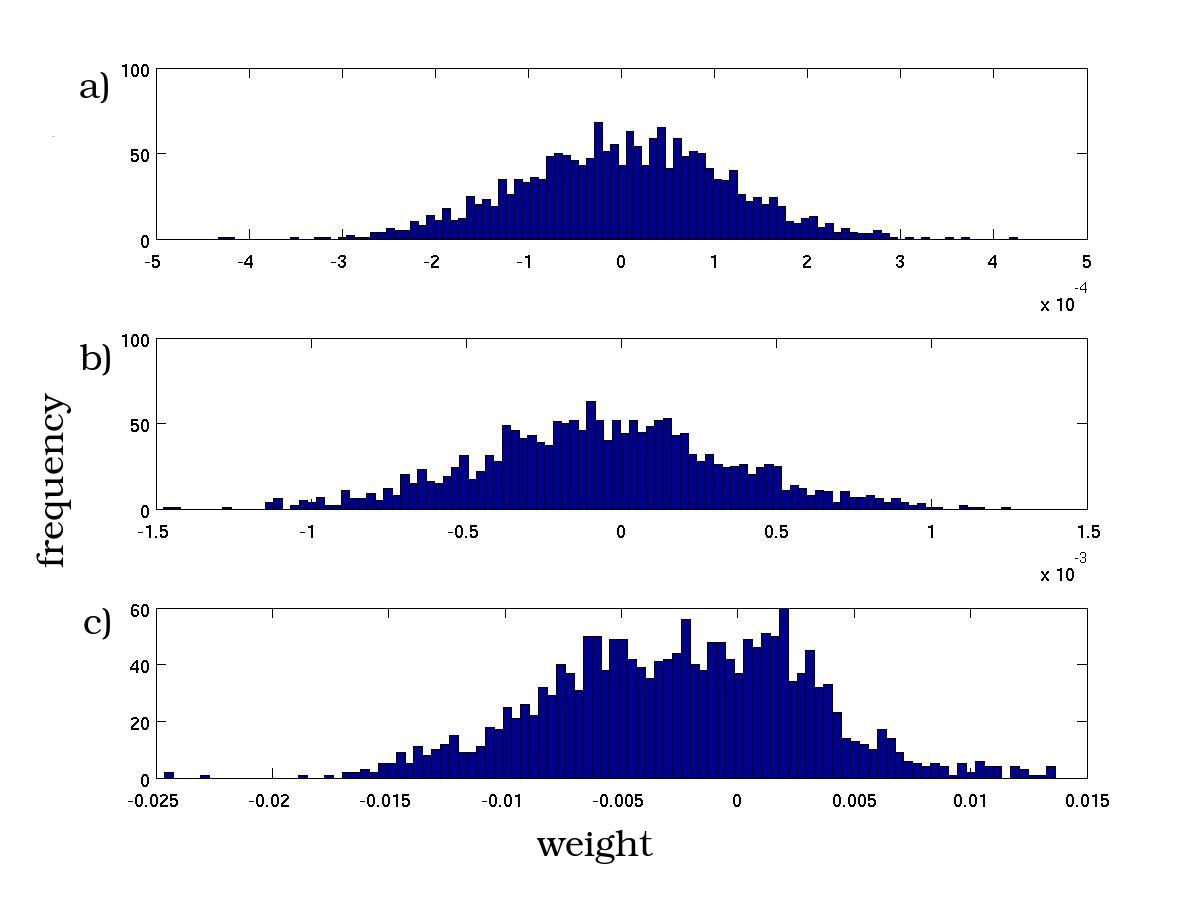}
\caption{Weight distribution of hidden-output connections at the initial phase (a), after the equilibration phase (b), and after the learning phase (c).}
\label{outp}
\end{center}
\end{figure}

We paid close attention to the weights of connections between input and hidden layers (see Fig. \ref{hidd}) and between hidden and output layers (see Fig. \ref{outp}), as well as to the weight distribution at the three stages of simulation: 

\begin{enumerate}
\item at the initial phase, which is reached after the mapping process;
\item after the network equilibration phase, when the weight deinforcement rule is applied to the network several times;
\item after the learning phase, when all the pattern pairs were successfully learned.
\end{enumerate}

The Gaussian distribution of weights can be found on the graphs at the initial phase of simulation. After equilibration is applied, the center of the distribution slightly shifts towards the left side for the connections between input and hidden layers and remains almost unaltered for the connections outgoing from the hidden layer.

A significant difference between the weight distribution dynamics can be observed among input-hidden and hidden-output connections after the training phase. In the first one, during the training phase, distribution expands over the range from -0.1 to 1 with a center at 0.45 and a significant concentration of weights around 0.5. However, the weights in the input-hidden connections shrink considerably from being in the range from -4 to 4 at the initial stage to the range from -0.02 to 0.015 at the final stage with an unclear distribution centered around -0.002. 

\begin{figure}
\begin{center}
\includegraphics[width=\columnwidth]{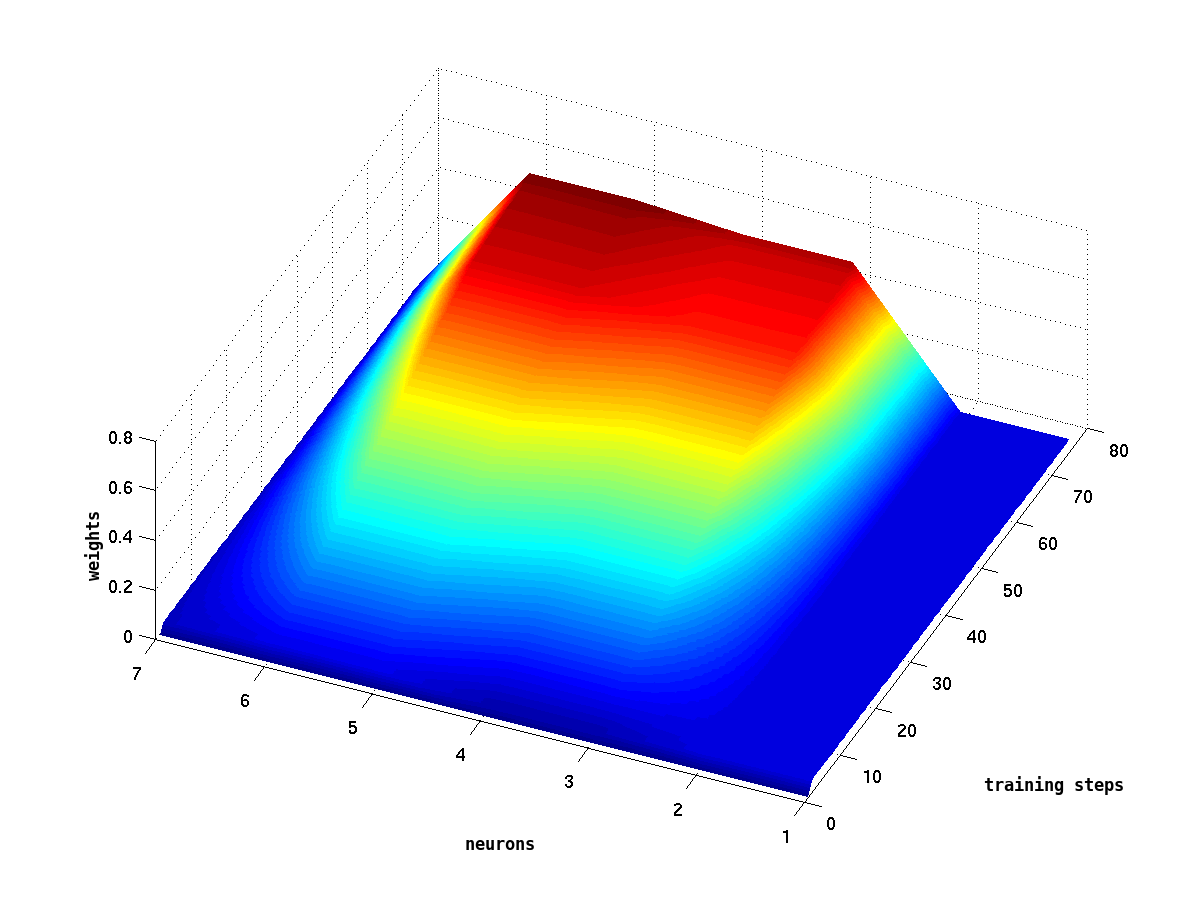}
\caption{Weight alteration of connections between input and hidden layers during the simulation progress (rotated for the best visibility).}
\label{w1}
\end{center}
\end{figure}

This phenomenon can be explained in the following way. The input neurons tend to excite only a specific set of hidden neurons and concentrate their connection efficacy to a limited number of neurons. Because of this, a change of the formed active paths requires more effort but makes the connections more stable and insensitive to changes. This behavior can be seen in Fig. \ref{w1}. 

\begin{figure}
\begin{center}
\includegraphics[width=\columnwidth]{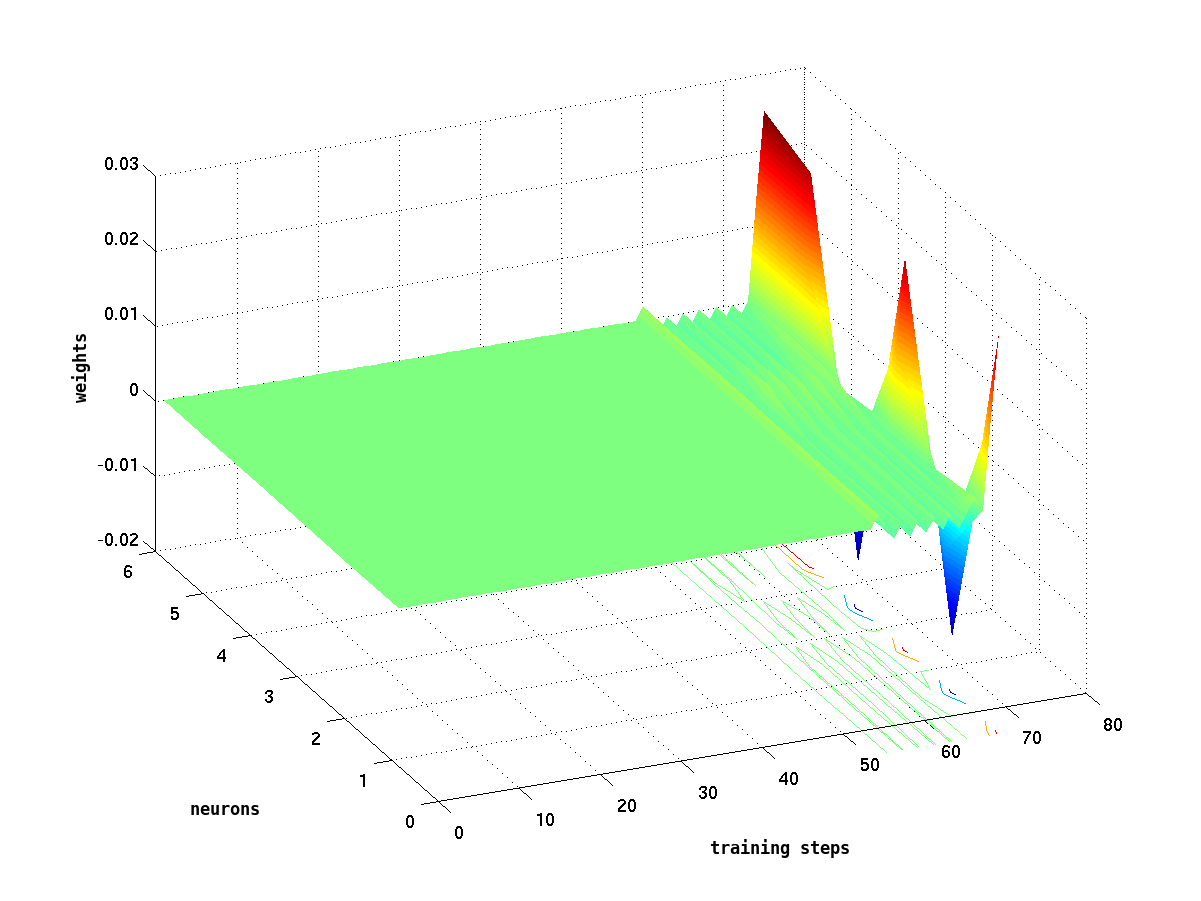}
\caption{Weight alteration of connections between hidden and output layers during the simulation progress (rotated for the best visibility).}
\label{w2}
\end{center}
\end{figure}

The hidden neurons broaden their influence on the wide range of neurons, although having a comparatively low influence on them. These connections are more sensitive to the input changes and faster adapt the required firing pattern by exciting the right output neurons. Fig. \ref{w2} shows that weights are almost unalterable during the initial phase of simulation, eventually changing their weights at the final simulation phase. 

The input-hidden connections adapt faster due to their lower number and only after their weights are settled, the hidden-output connections start actively adapting. The adaptation process is shown in  Fig. \ref{w1} and Fig. \ref{w2}, where it is clearly visible that the hidden-output weights start adapting. 

Concluding, there are two levels of learning in the three-layer neural network. Whereas the weights of the input neurons are less sensitive to changes, they better retain the learned patterns as opposed to the weights of the hidden neurons, which being highly-alterable and rapidly adapting, easily `forget' the learned patterns.

This allows us to assume that input-hidden connections are more influential to the adaption of the required output connection compared to the hidden-output connections. Taking into account that connections with large weights are highly sensitive (according to \cite{Chiu:1994}), their faulty behaviour is more critical for the proper operation of the network. Moreover, after the training phase is complete, a possible fault in the input-hidden connections is more devastating as it automatically leads to a faulty output in the corresponding post-synaptic nodes. 

\begin{figure}
\begin{center}
\includegraphics[width=\columnwidth]{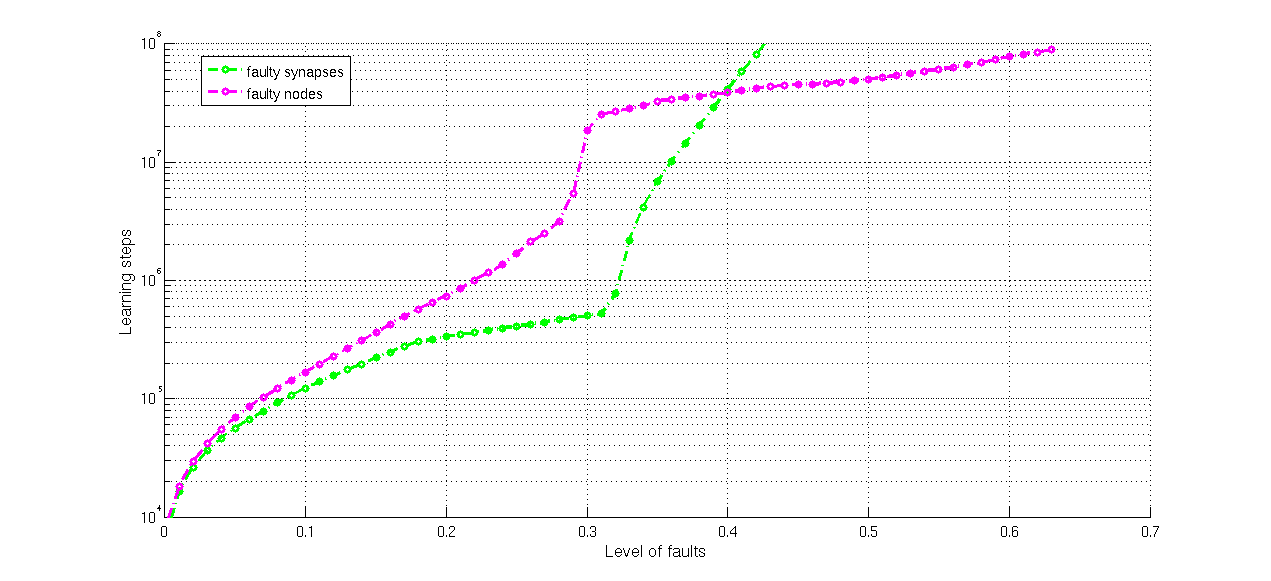}
\caption{Dependence of the number of learning steps required to learn 20 neurons by 500-neuron size network on the faults number injected at the network level during the learning process. The level of faults is presented in percents.}
\label{damage2}
\end{center}
\end{figure}

To assess the fault-tolerance of the neural network during the training stage, the fault injection technique was applied.  Fig. \ref{damage2} shows the dependence of the number of learning steps on faults. The red line represents the dependence on the faulty nodes and the green line represents the dependence on the faulty synapses. The level of faults is represented in percents rather than by the actual number of faults as this provides a better representation of the scale of damage done to the network. 

The dependence is almost linear up until 0.3\% for the network with damaged nodes and up until 0.25\% for the network with damaged synapses. The faulty nodes reduce the learning speed more significantly comparing to the faulty synapses when the number of faults is small. However, this changes to the opposite after the level of faults reaches 0.4\%, when the level of damage caused by faulty synapses becomes larger than by the faulty nodes. The value 0.4\% represent the point when the number of faulty synapses introduced to the network is equal to the number of faulty synapses caused by the faulty nodes.

\subsection{Fault Tolerance in Operation}
We assessed the fault-tolerance of the neural network 
during the operation phase. For this the fault injection technique was applied. In order to collect statistically valid data each simulation of the network of 2000 neurons (with 10 neurons dedicated to input and another 10 neurons to output, taught 20 pairs of patterns) was run 1000 times. Each time a certain number of faulty nodes was placed probabilistically according to the described fault injection technique and quality of the network output was measured by Eq. \ref{equation2}. The quality of output represents the global error and stands for the probability of receiving the expected output and is based on the maximum number of different bits between the expected patterns and the produced pattern. This experiment produced a plot of the neural network fault tolerance against the number of faulty nodes and synapses, by which the network reliability could then be judged.

\begin{figure}
\begin{center}
\includegraphics[width=\columnwidth]{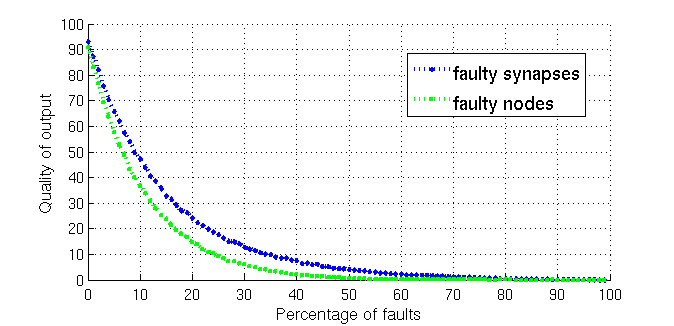}
\caption{The quality of output against the amount of faulty neural network's nodes while recalling pre-learned 20 patterns using the network of 2000 neurons.}
\label{FaultTolerance}
\end{center}

\begin{center}
\includegraphics[width=\columnwidth]{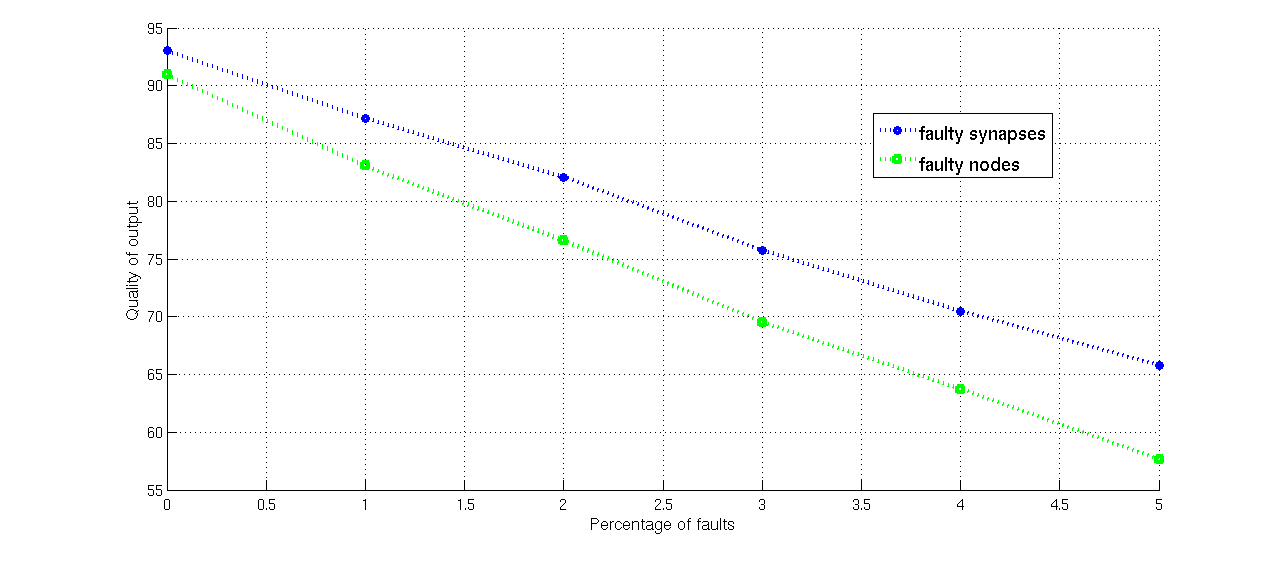}
\caption{The quality of output against the amount of faulty neural network's nodes while recalling pre-learned 20 patterns using the network of 2000 neurons (zoomed version of Fig. \ref{FaultTolerance}).}
\label{FaultTolerance2}
\end{center}
\end{figure}

Fig. \ref{FaultTolerance} shows the effect of faults injection while evaluating a set of 20 patterns. We presented the damage volume in percentage to the all possible damages in order to scale the amount of damage with the size of the physical implementation. The figure shows that performance degradation is in some sense graceful. According
to the plot, 5\% faulty nodes guarantees 60\% correct output and 10\% faulty nodes reduces the probability of the correct result to 50\%. A network with 2\% faulty nodes produces the correct result with a probability of 90\%. Fig. \ref{FaultTolerance2} presents the zoomed area of Fig. \ref{FaultTolerance}.

Any fault will influence the output to some degree since all components participate in any computation. This leads to graceful degradation being exhibited by most neural networks, i.e. neural networks will not suffer catastrophic failure, and also allows approaching failure to be detected by using a continuous reliability measure. The fault tolerance that results in this reliability is not inherent within neural networks: it does need to be specifically designed and built into them, and so the architectural complexity which often arises due to various fault tolerance techniques being used is absent in neural network systems. Finally, although any faults which do occur cannot be located, they can be removed from the system since neural networks can learn.

Although the experiments were performed on comparatively small networks (about 2000 neurons) with a small training set, the results are considered indicative for large networks with large training sets due to the scalable nature of calculations.

We conclude that fault tolerance arises because the computation is distributed  in the neural network rather than localized. Moreover, the system is self-organizing rather than being centrally configured.

\section{Conclusions}
Faults do and will occur in a system over time, and there always will come a time when performance is below acceptable limits. The issue of fault tolerance is of particular importance for the creation of large distributed machines based on communication-centric computation. However unlike traditional computing techniques, the neural network approach does not insist on exact computation. There is the strongly non-linear nature and the distribution of information or ``knowledge'' throughout all of the network. 

We presented a  distributed feed-forward neural network, which is insensitive to intermittent faults caused by unreliable communication or faulty hardware components. A review of work examining the fault tolerance of neural networks has been presented along with several techniques assuring fault-tolerant behavior. Also, various possible influences on the fault tolerance of neural networks were discussed. Among them, we show that uniform weight distribution offers the particular promise of more effective and faster training. We also show that reducing the connectivity between input and hidden layers is more advantageous prior to learning and the connections between hidden and output layers are less vulnerable to faults during learning and afterwards. 

Almost all of the units and connections participate in producing an output either directly or indirectly. Since it is difficult to exactly determine the required amount of processing units and their connections, their redundant number results in a higher degree of reliability. Thus the malfunctioning of a particular element of a system should not greatly affect the system's function if there is sufficient redundancy.


This analysis offers several more general lessons for building reliable systems on unreliable hardware. Clearly some redundancy is needed, and homogeneity is important, but self-organisation is also a necessary requirement.
\balance
\bibliographystyle{IEEEtran}
\bibliography{Journ_bibtex.bib}

\begin{thebibliography}{10}
\providecommand{\url}[1]{#1}
\csname url@samestyle\endcsname
\providecommand{\newblock}{\relax}
\providecommand{\bibinfo}[2]{#2}
\providecommand{\BIBentrySTDinterwordspacing}{\spaceskip=0pt\relax}
\providecommand{\BIBentryALTinterwordstretchfactor}{4}
\providecommand{\BIBentryALTinterwordspacing}{\spaceskip=\fontdimen2\font plus
\BIBentryALTinterwordstretchfactor\fontdimen3\font minus
  \fontdimen4\font\relax}
\providecommand{\BIBforeignlanguage}[2]{{%
\expandafter\ifx\csname l@#1\endcsname\relax
\typeout{** WARNING: IEEEtran.bst: No hyphenation pattern has been}%
\typeout{** loaded for the language `#1'. Using the pattern for}%
\typeout{** the default language instead.}%
\else
\language=\csname l@#1\endcsname
\fi
#2}}
\providecommand{\BIBdecl}{\relax}
\BIBdecl

\bibitem{top500}
J.~Dongarra, H.~Meuer, E.~Strohmaier, and H.~Simon, ``Top 500 supercomputer
  sites.'' May 2012, [Online]. Available: \url { http://www.top500.org/}.

\bibitem{Mukherjee:2003}
S.~S. Mukherjee, C.~Weaver, J.~Emer, S.~K. Reinhardt, and T.~Austin, ``A
  systematic methodology to compute the architectural vulnerability factors for
  a high-performance microprocessor,'' in \emph{Proc. MICRO 36}, 2003, pp.
  29--40.

\bibitem{SpiNNaker:2010}
S.~Furber and S.~Temple, ``Spinnaker a universal spiking neural network
  architecture,'' University of Manchester, Manchester, UK, Tech. Rep., Oct.
  2010.

\bibitem{SpiNNaker}
S.~Furber and D.~Lester, ``{SpiNNaker project}.'' May 2012, [Online].
  {A}vailable: \url { http://apt.cs.man.ac.uk/projects/SpiNNaker/}.

\bibitem{BlueBrain}
H.~Markram, R.~Bishop, and R.~Cicurel, ``{Blue Brain project}.'' May 2012,
  [Online]. Available: \url { http://bluebrain.epfl.ch/}.

\bibitem{Plana:2009}
J.~Navaridas, M.~Luj\'{a}n, J.~Miguel-Alonso, L.~A. Plana, and S.~Furber,
  ``Understanding the interconnection network of {SpiNNaker},'' in \emph{Proc.
  ICS}, 2009, pp. 286--295.

\bibitem{Moore:1989}
W.~R. Moore, ``Neural computers,'' R.~Eckmiller and C.~v.~d. Malsburg,
  Eds.\hskip 1em plus 0.5em minus 0.4em\relax New York, NY, USA:
  Springer-Verlag New York, Inc., 1989, ch. Conventional fault-tolerance and
  neural computers, pp. 29--37.

\bibitem{Mozer:1989}
M.~Mozer and P.~Smolensky, ``Skeletonization: A technique for trimming the fat
  from a network via relevance assessment,'' in \emph{Proc. NIPS}, D.~S.
  Touretzky, Ed.\hskip 1em plus 0.5em minus 0.4em\relax San Mateo: Morgan
  Kaufmann, 1988, pp. 107--115.

\bibitem{Tai:1990}
H.-M. Tai, ``Fault tolerance in neural networks,'' \emph{Proc. WNN-AIND},
  p.~59, Feb. 1990.

\bibitem{Belfore:1989}
L.~Belfore and B.~Johnson, ``The fault-tolerance of neural networks,''
  \emph{International Journal of Neural Networks Research and Applications},
  pp. 24--41, Jan. 1989.

\bibitem{Chiu:1993}
C.~Chiu, K.~Mehrotra, C.~Mohan, and S.~Ranka, ``Robustness of feedforward
  neural networks,'' \emph{Proc. ICNN}, vol.~2, pp. 783--788, March 1993.

\bibitem{Carter:1991}
M.~J. Carter and B.~Segee, ``Fault tolerance of pruned multilayer networks,''
  \emph{Proc. IJCNN}, vol.~2, pp. 447--452, July 1991.

\bibitem{Carter:1990}
M.~J. Carter, F.~J. Rudolph, and A.~J. Nucci, ``Advances in neural information
  processing systems 2,'' D.~S. Touretzky, Ed.\hskip 1em plus 0.5em minus
  0.4em\relax San Francisco, CA, USA: Morgan Kaufmann Publishers Inc., 1990,
  ch. Operational fault tolerance of CMAC networks, pp. 340--347.

\bibitem{Segee:1994}
B.~E. Segee and M.~J. Carter, ``Comparative fault tolerance of parallel
  distributed processing networks,'' \emph{IEEE Trans. Comput.}, vol.~43,
  no.~11, pp. 1323--1329, Nov. 1994.

\bibitem{Nijhuis:1990}
J.~Nijhuis, B.~Hofflinger, A.~van Schaik, and L.~Spaanenburg, ``{Limits to the
  fault-tolerance of a feedforward neural network with learning}.''\hskip 1em
  plus 0.5em minus 0.4em\relax IEEE Comput. Soc. Press, 1990, pp. 228--235.

\bibitem{Damper:1993}
M.~Emmerson and R.~Damper, ``Determining and improving the fault tolerance of
  multilayer perceptrons in a pattern-recognition application.'' \emph{IEEE
  Trans. Neural Netw.}, vol.~4, no.~5, pp. 788--93, Sept. 1993.

\bibitem{Phatak:1995}
D.~S. Phatak and I.~Koren, ``Complete and partial fault tolerance of
  feedforward neural nets,'' \emph{IEEE Trans. Neural Networks}, vol.~6, pp.
  446--456, 1995.

\bibitem{Seelen:1989}
W.~V. Seelen and H.~A. Mallot, ``Neural computers,'' R.~Eckmiller and C.~v.~d.
  Malsburg, Eds.\hskip 1em plus 0.5em minus 0.4em\relax New York, NY, USA:
  Springer-Verlag New York, Inc., 1989, ch. Parallelism and redundancy in
  neural networks, pp. 51--60.

\bibitem{Bolt:1991}
G.~Bolt, ``{Technical Report YCS 154: Investigating Fault Tolerance in
  Artificial Neural Networks},'' Dept. Computer Architecture Group, Universite
  of York, Tech. Rep., March 1991.

\bibitem{Panchal:2011}
G.~Panchal, A.~Ganatra, Y.~P. Kosta, and D.~Panchal, ``Behaviour analysis of
  multilayer perceptrons with multiple hidden neurons and hidden layers,'' in
  \emph{International Journal of Computer Theory and Engineering}, vol. 3(2),
  April 2011, pp. 332--337.

\bibitem{Huang:2003}
G.-B. Huang, ``Learning capability and storage capacity of two-hidden-layer
  feedforward networks.'' \emph{IEEE Trans. on Neural Netw.}, vol.~14, no.~2,
  pp. 274--281, March 2003.

\bibitem{Burr:1988}
D.~J. Burr, ``Experiments on neural net recognition of spoken and written
  text,'' \emph{IEEE Trans. Acoustics, Speech and Signal Processing}, vol.~36,
  no.~7, pp. 1162--1168, July 1988.

\bibitem{Bengio:2009}
Y.~Bengio, ``{Learning deep architectures for AI},'' Dept. IRO, Universite de
  Montreal, Quebec, Canada, Tech. Rep., 2007.

\bibitem{Bosman:2004}
R.~J.~C. Bosman, W.~A. van Leeuwen, and B.~Wemmenhove, ``Combining {H}ebbian
  and reinforcement learning in a minibrain model,'' \emph{Neural Netw.},
  vol.~17, no.~1, pp. 29--36, 2004.

\bibitem{Chiu:1994}
C.~Chiu, K.~Mehrotra, C.~K. Mohan, and S.~Ranka, ``Training techniques to
  obtain fault tolerant neural networks,'' in \emph{Proc. FTCS-24}, June 1994,
  pp. 360--369.

\bibitem{Kulakov:2011}
A.~Kulakov and M.~Zwoli\'{n}ski, ``{Reducing the active paths interference in
  the Chialvo-Bak minibrain model},'' in \emph{Proc. ICCMS}, vol.~2, Jan. 2011,
  pp. 677--681.

\end{thebibliography}

\end{document}